\documentclass[a4paper, 10pt, conference]{ieeeconf}      
\usepackage{FG2020}

\usepackage{tikz}
\usetikzlibrary{chains}
\usepackage{subcaption}
\usepackage{array}
\newcolumntype{C}[1]{>{\centering\let\newline\\\arraybackslash\hspace{0pt}}m{#1}}
\usepackage[bottom]{footmisc}
\usepackage{tikz}
\usepackage{verbatim}
\usepackage{amsmath}

 \usepackage{booktabs}
 \usepackage{nicefrac}
 
 \usepackage{multicol}
\usepackage{graphicx}
\usepackage{multirow}
\usepackage{tabularx}
\usepackage{xcolor}
\usepackage{comment}

\FGfinalcopy 

\IEEEoverridecommandlockouts                              
\def\FGPaperID{30} 

\title{\LARGE \bf
End-to-end facial and physiological model for \\Affective Computing and applications
}

\author{\parbox{16cm}{\centering
    {\large Joaquim Comas, Decky Aspandi and Xavier Binefa }\\
    {\normalsize
    Department of Information and Communication Technologies, Pompeu Fabra University, Barcelona, Spain}}
}

\begin{document}

\ifFGfinal
\thispagestyle{empty}
\pagestyle{empty}
\else
\author{Anonymous FG2020 submission\\ Paper ID \FGPaperID \\}
\pagestyle{plain}
\fi
\maketitle

\begin{abstract}

In recent years, affective computing and its applications have become a fast-growing research topic. Furthermore, the rise of deep learning has introduced significant improvements in the emotion recognition system compared to classical methods. In this work, we propose a multi-modal emotion recognition model based on deep learning techniques using the combination of peripheral physiological signals and facial expressions. Moreover, we present an improvement to proposed models by introducing latent features extracted from our internal Bio Auto-Encoder (BAE). Both models are trained and evaluated on AMIGOS datasets reporting valence, arousal, and emotion state classification. Finally, to demonstrate a possible medical application in affective computing using deep learning techniques, we applied the proposed method to the assessment of anxiety therapy. To this purpose, a reduced multi-modal database has been collected by recording facial expressions and peripheral signals such as electrocardiogram (ECG) and galvanic skin response (GSR) of each patient. Valence and arousal estimates were extracted using our proposed model across the duration of the therapy, with successful evaluation to the different emotional changes in the temporal domain.

\end{abstract}

\section{INTRODUCTION}

Emotions are essential factors in human behaviour which influence every social action \cite{Darwin, James, Cannon, Ekman_1}. The affective computing, which in its core originated from the efforts to understand these factors \cite{affective} has attracted considerable attentions lately due to its application on numerous venues, such as education \cite{e_learning}, healthcare \cite{Liu}, etc.

In literature, a common approach to infer emotion states is by utilizing several modalities such as expressions, speech, body gestures, physiological signals, etc. Though the facial expressions are gaining more popularity due to their intuitive nature \cite{action_units,facial,tensorflow}, the physiological signals offer unique advantages to other modalities. First is their growing availability supplemented by recent arise of wearable devices usages. Secondly, they are quite invariant against external visual noises, such as illumination therefore are quite robust and versatile. Third advantage includes their fidelity qualities, since it is very hard to replicate or to mask these signals to simulate specific emotions. Lastly, they have relatively low dimensional structure allowing more efficient processing.

Nowadays, the emergence of large dataset such as AMIGOS \cite{AMIGOS} has opened a new possibility to use powerful deep neural networks to this field of affective computing \cite{tensorflow}, \cite{face}. However, compared to other computing fields, the adoption of deep learning techniques to process these physiological signals is still sparse \cite{towards_bio}, with the only the recent works of \cite{signals_0}, \cite{signals} are being the exceptions. Most studies use the raw signal as their input to their models with the assumption that they able to learn relevant features for their estimations directly \cite{auto_lstm, towards_bio}. However, we argue that this approach may lead to sub-optimal results since raw signals prone to contain a substantial amount of noises \cite{bioNoise1,bioNoise2}. This problem can be circumvented by the efficient use of auto-encoder \cite{autoencoder}, which is able to construct these relevant features in a compact way. Thus, we may have a less noisy features. This has been shown to improve the model estimates on other computing fields, such as facial recognition \cite{aefacerecog}, object classification \cite{aeobjectclas}, music generation \cite{aemusic}, data compression \cite{lossy_image}, dimensionality reduction \cite{dimensionality} etc. Nonetheless, in the affective computing, it remains largely unexplored, with the only examples are the works of Tang et al. \cite{auto_lstm} and Yildirim et al. \cite{ecg_auto}. Unlike these approaches, however, our models fully differentiate the stream input from bio-signal and facial features, due to the large size of each modality which we argue will require specific processing. We realize this by adopting single modality auto-encoder to achieve compact representation of bio-signal \cite{ecg_auto} that simultaneously allows us to quantify its individual contribution on improving the quality of models estimates, which currently is still largely unexplored.




In this work, we propose end-to-end models for automatic affect estimations using multiple modalities. In conjunction, we also introduce new dataset collected from patients who have been exposed to anxiety treatments with aim to expand the application of affective computation. Different to previous approaches, we capitalize on the individual use of latent features extracted from our internal Bio Auto-Encoder alongside stream of spatial features to improve the accuracy of our models estimates. Using recently published AMIGOS \cite{AMIGOS} dataset, we will perform an analysis to reveal the effectiveness of each modality. Then, the combination of extracted latent features will be used to improve our model estimates. Next, we will present a relative comparison of our results against related studies. Lastly, we will demonstrate a real-world application of our trained models by evaluating our model estimates using recorded patient data, collected before and after treatment. Specifically, the contributions of our work are as follow :

\begin{enumerate}
    \item We propose improvements by incorporating of latent features extracted by means of our internal Bio Auto-Encoder to our bio multi-modal network. 
    \item We present a new dataset collection from a medical therapy to expand the application of affective computing model.
    \item We perform a thorough analysis to confirm the benefit of the utilisation of multi-modal inputs, which will be complemented by the correspondent latent features.
    \item We present our competitive results against other related studies on AMIGOS \cite{AMIGOS} dataset and its real-world capability of our models to estimate the patient emotion state during a therapy.
    
\end{enumerate}



\section{Related work}

One of the earliest use of physiological data for automatic emotion estimation has been in the work of Fridlund et al. \cite{Izard}, where they applied linear discriminant analysis on the facial electromyographic (EMG) activity. They reported that there is a correlation between personal biological signal and their emotional activity. A multi-modal approach was introduced by Picard et al. \cite{Picard}, combining four separate modalities: heart rate, skin conductance, respiration and facial electromyogram. Using this more extensive data, they managed to achieve relatively better results.

Other research attempt to concentrate on finding the correlation between emotion and physiological signal, such as the work of Nasoz et al. \cite{Nasoz} and Feng et al \cite{Feng}. Nasoz et al. proposed an experiment by eliciting certain types of emotions contained in the movie clip presented to the participant. They utilised k-nearest neighbour, discriminant function analysis and Marquardt back-propagation algorithm, using features from several modalities: galvanic skin response, temperature and heart rate. While Feng et al. adopted support vector machines (SVM) classifier to perform an automated method for emotion classification using EDA signals with wavelet-based features. In other hand, Soleymani et al. \cite{Pun} introduced new modality of electroencephalography (EEG), pupillary response and gaze distance with decision level fusion (DLF) technique to produce more accurate results on each modality than previous methods.

Due to the lack of huge data to facilitate extensive comparisons, Koelstra et al. created DEAP dataset \cite{DEAP} which offers large number of response features such as electrocardiogram (ECG), galvanic skin response (GSR) or electrodermal activity (EDA), electrooculargram (EOG) and electromyogram (EMG) to expand the possibility to analyse such responses. Using this dataset, Tang et al. \cite{auto_lstm} further introduced bimodal deep denoising autoencoders to extract high level representations of both bio-signal and visual information with bimodal LSTM as bottleneck to analyze the impact of additonal temporal information. This possibility to construct the latent features from bio-signal has been studied further on the work of Yildirim et al. \cite{ecg_auto} which introduces efficient compression on similar ECG data. Albeit still to date, none of these approaches fully analyze and quantify the specific importance of the extracted bio-latent features on the final model estimations.


Currently, the biggest dataset so far that allows the analysis of the effect of physiological signals in emotion recognition is the AMIGOS dataset \cite{AMIGOS}. This dataset enables more extensive investigations of specific peripheral signals of ECG and GSR, and has been used on multiple literatures. First example is the work of Gjoreski et al. \cite{interdomain} who presented an inter-domain study for arousal recognition using RR-intervals and GSR. Other works tried to adopt variational auto-encoders network(VAE) to learn personality-invariant physiological signal representation by Yang et al. \cite{vae} with improved accuracy. Another study is the work of Santamaria et al. \cite{Amigos_1}, where they introduced deep convolutional neural networks largely outperform previous results. Finally, Siddarth et al. \cite{towards_bio} applied a novel deep learning method to different physiological and video data, and currently holdings state of the art accuracy on this dataset. Although these models have been shown to work well in their specific domains, there still has not been any investigation to asses their capability on real-life settings, such as in medical application.

\section{Methodology}

In this section, we will explain datasets used in our experiments, including our data pre-processing steps. Further, we describe our proposed multi-modal network, which operates by incorporating existing bio-physiological responses and facial features for affect estimations. 



\subsection{Datasets and pre-processing}

\subsubsection{AMIGOS Dataset}

 AMIGOS database \cite{AMIGOS}, which stands for mood, personality and affect research on individuals and groups, was collected using two different experimental settings. The first setting involved 40 participants watching 16 short videos (trial length varying between 51 and 150 seconds). In the second part, some participants watched four long videos on different scenarios, i.e. individually and in groups. During these experiments, EEG, ECG and GSR signals were recorded using wearable sensors (all signals pre-processed at 128 Hz), while face and depth data were recorded by separate types of equipment. 
 
\begin{figure}[h!]
\centering
    \includegraphics[width= 85mm,scale=1]{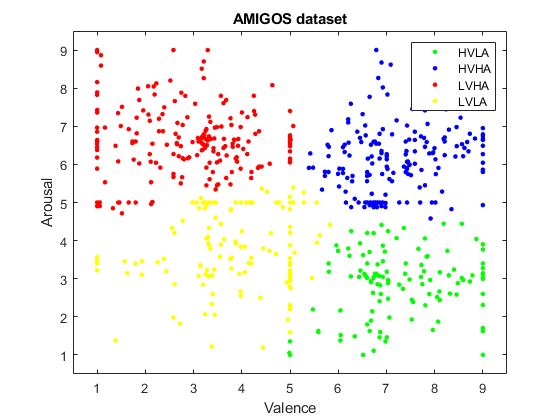}
    \caption{Distribution of valence and arousal label from self-assessment labels of AMIGOS dataset (scale from 1 to 9) classified in four quadrants (HVLA: high-valence / low-arousal, HVHA: high-valence / high-arousal, LVHA: low-valence / high-arousal), LVLA: low-valence / high-arousal}.
    \label{fig:amigos_anali}
\end{figure}

 In total, there are 640 instances (16 trials x 40 participants) along with their respective tag for valence, arousal, liking, familiarity and seven basic emotions (neutral, disgust, happiness, surprise, anger, fear, and sadness) available. The first four affect dimensions are measured in a continuous scale of 1 to 9 with basic emotions represented using one-hot label. Figure~\ref{fig:amigos_anali} visualises the distribution of the self-assessment label in terms of valence and arousal of the dataset. Notice that most of the emotion examples are located around the centre of each quadrant, which are relatively close to the neutral emotion state.
 

 
 
 
 

\subsubsection{Medical Therapy Dataset}

In this work, we introduce a new multi-modal data collection from several patients who have undergone an anxiety treatment. Using this data, we would like to apply and provide more objective analysis based on the patient bio-responses while being subjected to a complementary polarisation treatment \cite{therapy}. In this case, our hypothesis is that there will be changes in terms of valence and arousal over the therapy.  

\begin{figure}[h!]
\centering
\includegraphics[width=.15\textwidth, height=2.5cm]{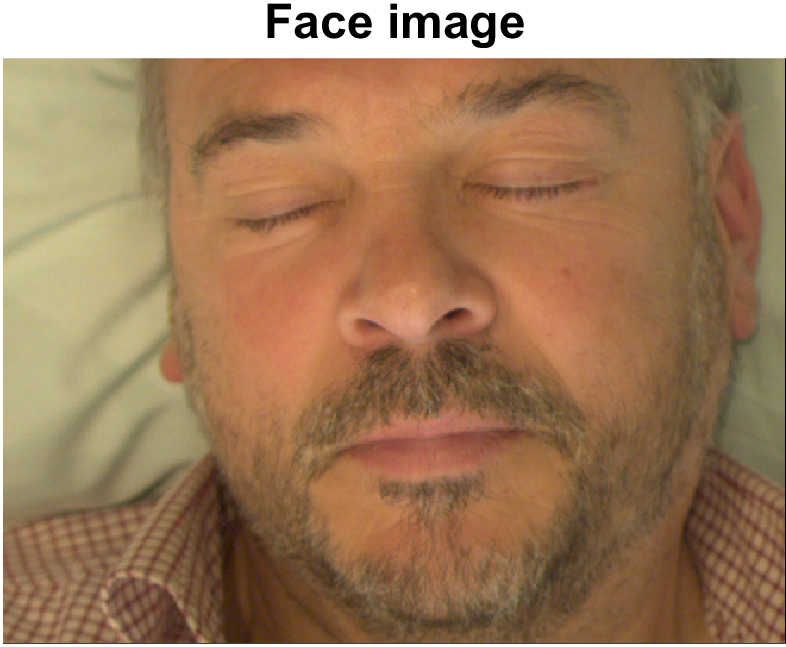}\\
\includegraphics[width=.35\textwidth,  height=2cm]{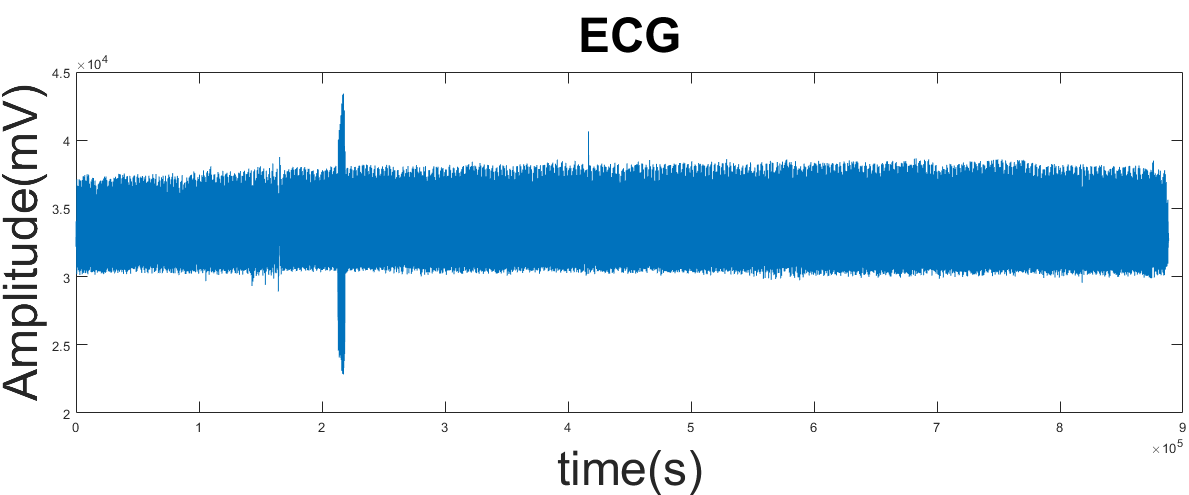}\\
\includegraphics[width=.35\textwidth, height=2cm]{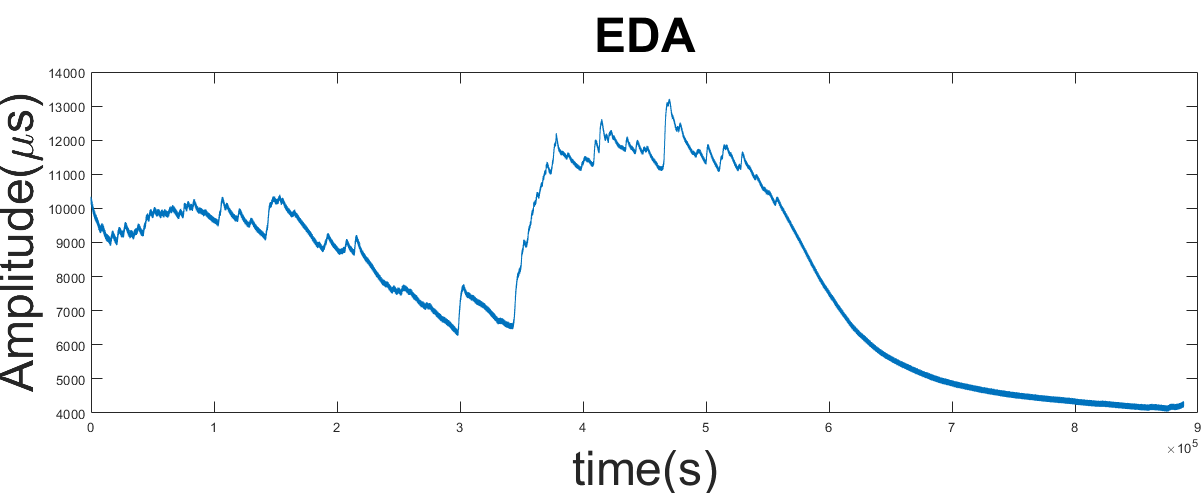}\\
\caption{Example of data collection of a patient during therapy}
\label{fig:collected_data}
\end{figure}

During the therapy session, we recorded several physiological signals using wearable sensors along with their face in a synchronous way. The recordings were performed with a sampling rate of 800 Hz for bio-signals collection, while 60 frames per second for the facial images. An example of data collection of a patient during the therapy session can be seen in Figure~\ref{fig:collected_data}. We collected self-assessment questionnaires from each patient prior and after treatment that provide their subjective perceptions of therapy. Currently, we have successfully recorded a sample from five patients during a single therapy session, where we expect to gather more in our future works. This sample consists of three females and two males with the mean age of 42.8 years and 10.40 years of standard deviation. 


To perform the treatment validation, we adopt the four-quadrant circumplex model \cite{Russell} by combining the value of both valence and arousal of emotional states. In the ideal scenario, there should be a movement tendency of the patients emotional state across quadrants. Initially, the emotional state will be located around the second quadrant (low valence and positive arousal) indicating high-level stress of the patients. After the treatment, we expect its location to be around fourth quadrant, i.e. high valence and low arousal showing the patients tranquillity.




\subsubsection{Data Preprocessing}
To ease our model training and estimations, we performed multiple stages of data pre-processing to each modality used by our model : facial features, electrocardiogram (ECG) and electrodermal activities (EDA). The first stage was to locate the facial area to enable us to remove its respective backgrounds. Since they may contain other redundant parts of the scene which consequently slow down the training process. To do this, we used facial tracking model of \cite{landmarks} and cropped facial area given detected facial landmark. 




The second stage involves scaling the physiological signals, which will be crucial to obtain a better reconstructed solution in the auto-encoder network.  Considering that data is already down-sampled, pre-processed and segmented, no more signal pre-processing was needed.


The last stage was to maintain the temporal relation of the physiological signals with facial images through data synchronisation. Specifically, for each frame, we segmented each bio-signals with a length of 1000 samples, that corresponds to eight heartbeats. We also performed padding to the initial and end of each bio-signal to accommodate this scheme. 



    
\subsection{Bio Multi-Modal with Internal Auto-Encoder Network}
\begin{figure*}[h!]
    \centering
    \includegraphics[width=167mm,scale=0.60]{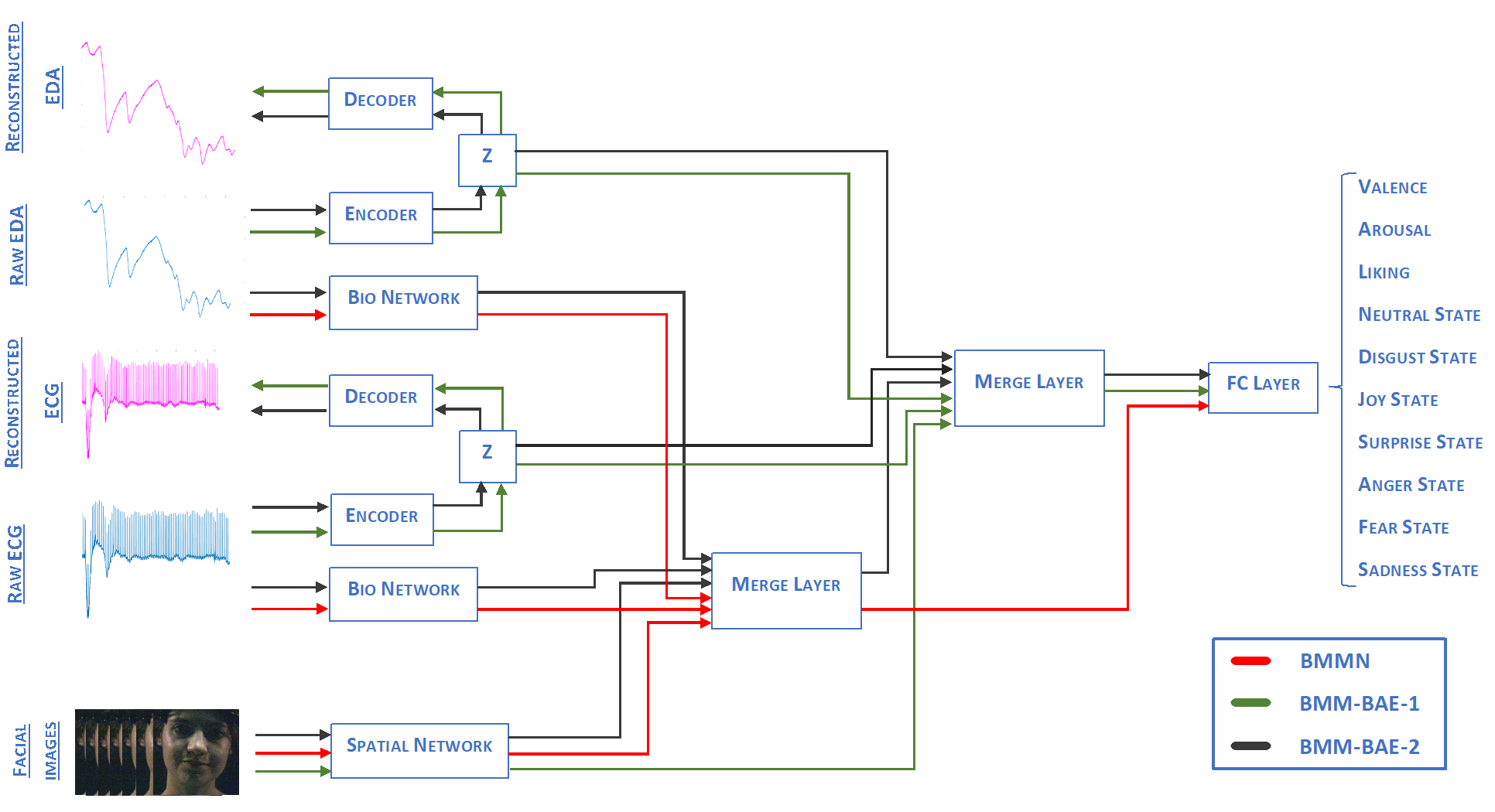}
    \caption{ Overall structure of our proposed Bio Multi-Modal with Internal Auto-Encoder Network(BMMN). Red line shows the pipeline of our base network of BMMN, green line for BMMN-BAE-1 and black line for BMMN-BAE-2}
    \label{fig:multi_baseline}
\end{figure*}

Our proposed model of Bio Multi-Modal Network (\textbf{BMMN}) operates by involving several modalities to estimate the people affect states while fully end-to-end trainable. BMMN model serves as the baseline of our results, which we will later improve by the incorporation of latent features captured by our Bio Auto-Encoder (\textbf{BAE}). Figure~\ref{fig:multi_baseline} depicts the whole structure of our proposed models.  

\subsubsection{Bio Multi-Modal Network}


Our BMMN consists of three main parts: Bio Network, Spatial Network, and merging layer. Bio Network is dedicated to compute bio features given the raw physiological signals, which in this case are ECG and EDA. On the other side, the spatial network creates visual features given the cropped face. These features then merged on the Merge Layer and passed to the final bottleneck fully connected layer (FC). FC layer in the end produces the affect estimates: valence, arousal, liking and seven emotional states (neutral, disgust, joy, surprise, anger, fear, and sadness).  

We construct our Bio Network by using stacked of 1D convolutional layers which later joined step-wise through residual connections \cite{he2016deep}. Specifically, we propose to utilise different kernel layers size, which has been shown to learn more efficient representations using their sparse filter relationship \cite{inception}. Furthermore, we also introduce a skip connection to previous outputs to ease the gradient while training \cite{he2016deep}. The full network architectures of Bio Network is summarised in Table~\ref{table_signals}. As for our Spatial Network, we rely on pre-trained ResNet-50 \footnote{http://pytorch.org/docs/stable/torchvision/models.html} which is available on the standard PyTorch-torchvision \cite{pytorch} libraries by performing transfer learning.

\begin{table}[h!]
\renewcommand{\arraystretch}{1.12}
\centering
\begin{tabular}{C{1.18cm} C{1 cm} C{1.18cm} C{0.75cm} C{0.75cm} C{1.1cm}} 
 \hline
 Layer & Kernel size & Activation & Filters No. & Stride & Output size\\ [0.1ex] 
 \hline
  Input & - & - & - & - & 1000 x 1  \\ [0.5ex]
  Conv1 & 200 x 1 & ReLU & 4 & 1 & 805 x 4 \\ [0.5ex]
  Maxpool1 & 2 x 1 &  - & 4 & 2 & 402 x 4 \\ [0.5ex]
  Conv2 & 100 x 1 & ReLU & 2 & 1 & 307 x 2 \\ [0.5ex]
  Maxpool2 & 2 x 1 &  - & 2 & 2 & 153 x 2 \\ [0.5ex]
  Conv3 & 50 x 1 & ReLU & 2 & 1 & 357 x 2 \\ [0.5ex]
  Maxpool3 & 2 x 1 &  - & 2 & 2 & 178 x 2 \\ [0.5ex]
  Conv4 & 25 x 1 & ReLU & 2 & 1 & 382 x 2 \\ [0.5ex]
  Maxpool4 & 2 x 1 &  - & 2 & 2 & 191 x 2 \\ [0.5ex]
  Merge & - & ReLU & - & - & 2652 x 1 \\ [0.5ex]
  Output & - & - & - & - & 2652 x 1  \\ [0.5ex]
 \hline
 \end{tabular}
\caption{Detailed architecture of the proposed 1D-CNN for Bio Network.}\label{table_signals}
\end{table}





\subsubsection{Bio Auto-Encoder Network}
Bio Auto-Encoder Network (BAE) is responsible to extract a fixed length of latent features that represent the overall bio-physio signal input it received. We hypothesise that utilizing this vector representation can ease the model learning, and thus improve global estimations. We use similar stack of 1D convolutions used on Bio Network layer but arranged to follow standard encoder and decoder structure. We use max-pooling layers on the encoder part to reduce the dimension of the input features. With an intermediate fully connected layer (FC) to obtain a fixed length of latent features of \emph{z}. Subsequently by using the \emph{z} features, we get the reconstructed signal back with series of un-pooling layers with ReLU activation. The detail of layers used in BAE network can be seen on Table~\ref{table_auto}.





BAE network ultimately will generate compact \emph{z} features of 128 length vectors, in which we consider its representation quality based on the reconstructed output. The degree of their similarity indicates the robustness of extracted features \emph{z}, i.e. more similar means more robust the extracted features. This is due to inherent structures of auto-encoder which forces the inner hidden layers to discover more relevant features to reconstruct original signal.

\begin{table}[h!]
\centering
\renewcommand{\arraystretch}{1.04}
\centering
 \begin{tabular}{C{1.18cm} C{1 cm} C{1.18cm} C{0.75cm} C{0.75cm} C{1.1cm}} 
 \hline
 Layer & Kernel size & Activation & No. of filters & Stride & Output size\\ [0.1ex] 
 \hline
  Input & - & - & - & - & 1000 x 1  \\ [0.5ex]
  Conv1 & 200 x 1 & ReLU & 16 & 1 & 801 x 16 \\ [0.5ex]
  Maxpool1 & 2 x 1 &  - & 16 & 1 & 402 x 16 \\ [0.5ex]
  Conv2 & 100 x 1 & ReLU & 8 & 1 & 301 x 8 \\ [0.5ex]
  Maxpool2 & 2 x 1 &  - & 8 & 1 & 150 x 8 \\ [0.5ex]
  Conv3 & 50 x 1 & ReLU & 4 & 1 & 101 x 4 \\ [0.5ex]
  Maxpool3 & 2 x 1 &  - & 4 & 1 & 50 x 4 \\ [0.5ex]
  FC &  - &  - & - & - & 128 x 1 \\ [0.5ex]
  Unpool1 & 2 x 1 &  - & 4 & 1 & 101 x 4 \\ [0.5ex]
  Conv4 & 50 x 1 & ReLU & 8 & 1 & 150 x 8 \\ [0.5ex]
  Unpool2 & 2 x 1 &  - & 8 & 1 & 301 x 8 \\ [0.5ex]
  Conv5 & 100 x 1 & ReLU & 16 & 1 & 400 x 16 \\ [0.5ex]
  Unpool3 & 2 x 1 &  - & 16 & 1 & 801 x 16 \\ [0.5ex]
  Conv6 & 200 x 1 & ReLU & 1 & 1 & 1000 x 1 \\ [0.5ex]
  Output & - & - & - & - & 1000 x 1  \\ [0.5ex]
 \hline
 \end{tabular}
\caption{Detailed architecture of the Bio Auto Encoder Network.}\label{table_auto}
\end{table}







\begin{figure*}[h!]
\centering
\includegraphics[width=.24\textwidth,height=3.5cm]{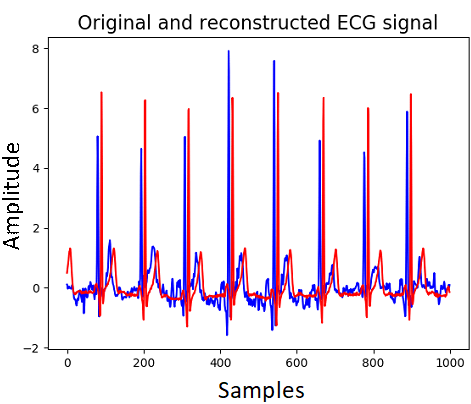}\hfill%
\includegraphics[width=.24\textwidth,height=3.5cm]{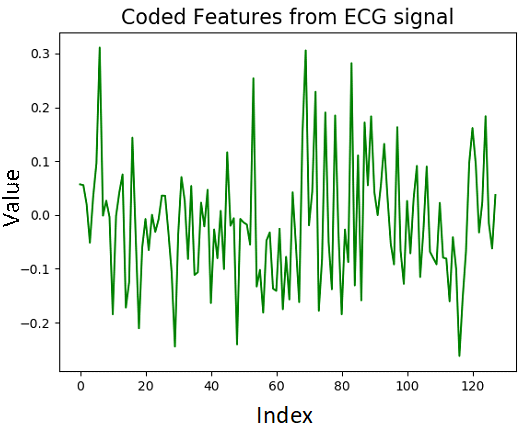}\hfill%
\includegraphics[width=.24\textwidth,height=3.5cm]{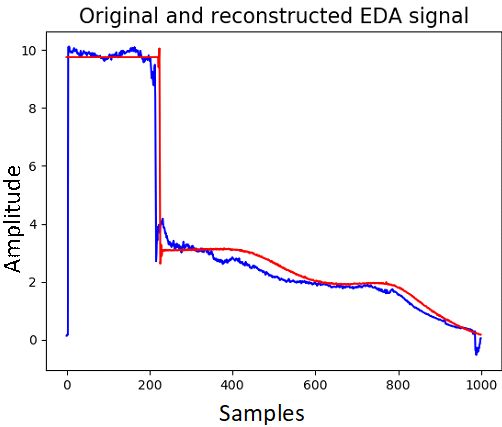}\hfill%
\includegraphics[width=.24\textwidth,height=3.5cm]{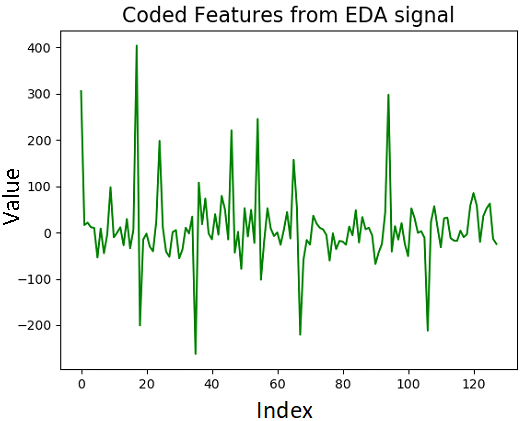}\hfill%
   
\caption{Reconstructed signals (red) overlaid against the original signals (blue), followed by respective latent features (green) for ECG and EDA sequentially.} 
\label{fig:autoencoder}
\end{figure*}

\subsection{Feature Combinations}
\label{sec:featureCombine}

We propose two combination strategies to improve the estimation of our baseline models of BMMN by involving the latent features \emph{z} generated by BAE: 
\begin{enumerate}
    \item The first strategy is to solely use z vector in place of features from Bio Network, and passing them to merge layer further for final estimation process. Since we assume original data could contain noisy or redundant information which can affect the emotion recognition process and a more compact feature of \emph{z} alone is sufficient to provide relevant information. This combination produces our second network of \textbf{BMMN-BAE-1}. 
    \item The second approach involves combining latent \emph{z} as auxiliary features along the essential features from the Bio Network, which later merged to follow further affect estimation pipelines. This is based on our assumption that latent features \emph{z} can serve as complementary information, thus enriching the input features which may ease the model estimations as well as improving final accuracy. We name our third alternative network as \textbf{BMMN-BAE-2}.
\end{enumerate}
Both of these network variants and their combination schemes can be seen on Figure \ref{fig:multi_baseline}.


\subsection{Model Loss and Training Setup}

We initially trained each network separately to ease the overall training processes, which we later perform joint training of all involving sub-networks to obtain our final estimations. We first trained BMMN for affect estimations to get our baseline results using standard $\ell^2$ loss. In this phase, we also trained BMMNs using each modality separately, i.e. to use either facial feature or physiological signal only. This to provide analysis for each modality described in Ablation Analysis (Section \ref{sec:ablation}).  

Secondly, we pre-trained BAE to reconstruct the input signals to capture relevant features prior integration so that it has been conditioned enough to the characteristics of these signals. Figure~\ref{fig:autoencoder} shows the reconstructed signals overlaid on top of the original bio-signal input, followed by its respective 128 coded latent features. As we can see, the reconstructed input signals are quite similar to original signals suggesting that our BAE able to construct corresponding latent features.

Thirdly, we combined the \emph{z} features into BMMN network following the scheme explained on Section \ref{sec:featureCombine} and further jointly trained them. The total loss for both BMMN-BAE-1 and BMMN-BAE-2 is as follow:

\begin{equation}
    L_{Total} = \lambda_{BMMN} L_{BMMN} + \lambda_{Bae} L_{Bae}
\end{equation}
where $L_{Bae}$ denotes the reconstruction loss between original and reconstructed signal and original signal, $L_{BMMN}$ is the loss for emotion label estimatons, $\lambda$ signifies the regularizer coefficients \cite{MultiTask} and the $L$ stands for standard $\ell^2$ loss.

On all training instances, we used a uniform distribution for weights initialization and Adam optimizer with a learning rate of 0.0001 for optimization. Using single NVIDIA Titan X we were able to use a batch size of 115 and took approximately two days to train each individual model. All models were implemented using the Pytorch framework \cite{pytorch}.  

    
\section{Experiments}
In this section, we will provide the results of our models using both pre-processed dataset of AMIGOS \cite{AMIGOS} and therapy dataset. With AMIGOS dataset, we first provide the analysis to see the impact of each modality used by our model to its final affect estimates, including the use of internally extracted latent features. Secondly, to see the quality of affect estimation from our proposed approach, we will compare our results with the reported results from related literature on this dataset. Then, we use our best performing model to validate the emotion state of each individual on our collected dataset. To evaluate the quality of affect estimates, we follow the original person-independence protocols of AMIGOS dataset. Finally, precision score (in per cent unit) is used to judge the quality of each estimated affect labels.


\subsection{Ablation Studies}
\label{sec:ablation}
We provide two principal analysis in these ablation studies: modality and latent feature impact analysis. Modality analysis exemplifies the impact of individual modalities as input to our BMMN and to establish our baseline accuracy. While latent feature impact analysis substantiates the benefit of incorporating hidden features \emph{z} for more accurate affect estimates of our joint models explained on section \ref{sec:featureCombine}.

\subsubsection{Modality Analysis}
To see individual contribution of each modality on the final estimates of our BMMN network, we trained BMMN for each modality separately by removing one modality to the other. This results in two other trained BMMN networks, with one, was trained only on physiological signals, and the other was using facial features only. We then compared their results against the normal BMMN, which received both bio-physio and facial features for a more complete analysis.  

\begin{table}[h!]
\renewcommand{\arraystretch}{1.3}
\centering
  \begin{tabular}{c|c c c}
    \hline
    \multirow{2}{1cm}{ Label} & \multicolumn{3}{c}{Modality}\\
    
    \cline{2-4}
    & Bio-signals & Faces & Multi-Modal\\
    \hline
    Valence & 58,25 \% & 56,67 \%&	\textbf{67,91 \%} \\ [0.65ex]
    
    Arousal & 55,65 \%& 78,28 \%& \textbf{78,36 \%}\\ [0.65ex]
    
    Liking & 69,01 \%& 75,16 \%& \textbf{77,82 \%}\\ [0.65ex]
    
    Neutral & 40,81 \%& 36,25 \%& \textbf{48,91 \%}\\ [0.65ex]
    
    Disgust & 55,08 \%& 26,01 \%& \textbf{75,92 \%}\\ [0.65ex]
    
    Joy & 45,88	\%& 40,48 \%& \textbf{70,27 \%}\\ [0.65ex]
    
    Surprise & 46,11 \%& 32,70 \%& \textbf{77,27 \%}\\ [0.65ex]
    
    Anger & 35,69 \%& 36,53 \%& \textbf{51,18 \%}\\ [0.65ex]
    
    Fear &  30,84 \%& 36,31 \%& \textbf{62,09 \%}\\ [0.65ex]
    
    Sadness & 29,88 \%& 26,24 \%& \textbf{65,55 \%}\\ [0.65ex]
    \hline
    
    Average & 46,92  \%& 44,46 \%& \textbf{67,53 \%}\\ [0.65ex]
    
    \hline
  \end{tabular}
  \caption{Results of BMMN utilizing bio-signals only, faces only, and both bio-signals and faces (multi-modal)}\label{baseline_modality}
\end{table}


Table~\ref{baseline_modality} provides results of trained models given specific modality of bio-signals only, faces only, and both (multi-modal). Based on these results, we can see that utilising only bio-signals and faces separately lead to comparable results. Mainly, bio-signals produced more stable accuracy across affects labels with 46,92 \% total mean of accuracy. While using facial images results in similar, but unstable estimates with higher accuracy in some affect labels, such as arousal and liking while lower in other, with total accuracy of 44,46 \%. This instability may the results of high parameter contained on our Spatial Networks. In general, these results are inferior when compared to our standard BMMN, i.e uses both modalities, with average accuracy of 67.53\%. We can observe that in overall, it produces higher accuracy across all emotion with close to 10 \% improvement on the valence estimates, indicating that bio-signals modality helps in classifying this particular affect dimension.

These findings suggest that both faces and bio-signals gives equal contribution to our BMMN models. However, we also note that our internal Spatial Network of BMMN has been externally pre-trained on other datasets, which shows the effectiveness of the structure of our Bio Network to be able to achieve comparable results. Furthermore, by utilising both modalities arranged in multi-task ways, we were able to improve overall affect estimates of our BMMN model outperforming their results when used separately.

\subsubsection{Impact of Latent Feature Analysis}

In this part, we present our baseline results from previous sections (BMMN) versus our other two model variants: BMMN-BAE-1 and BMMN-BAE-2. Table ~\ref{table:final results} shows the overall comparisons of these models. 

\begin{table}[h!]
\renewcommand{\arraystretch}{1.3}
\centering
  \begin{tabular}{l|c c c}
    \hline
    \multirow{2}{1cm}{ Label} & \multicolumn{3}{c}{Models}\\
    
    \cline{2-4}
    & BMMN & BMMN-BAE-1 & BMMN-BAE-2\\
    \hline
    Valence  & \textbf{67,91 \%}  & 66,19 \%& 65,05 \%\\ [0.75ex]
    Arousal  & 78,36 \%& 84,86 \%& \textbf{87,53 \%} \\ [0.75ex]
    Liking  & 77,82 \% & \textbf{78,76 \%} & 78,10 \% \\ [0.75ex]
    Neutral  & 48,91 \% & 44,29 \%& \textbf{53,68 \%}\\ [0.75ex]
    Disgust  & 75,92 \% & 76,86 \%& \textbf{92,62 \%} \\ [0.75ex]
    Joy  & \textbf{70,27 \%}  & 68,86 \% & 68,30 \%\\ [0.75ex]
    Surprise & \textbf{77,27 \%}  & 73,90 \% & 72,94 \%\\ [0.75ex]
    Anger  & 51,18 \% & 55,62 \%& \textbf{62,25 \%}\\ [0.75ex]
    Fear & 62,09 \% & \textbf{68,86 \%} & 65,15 \%\\ [0.75ex]
    Sadness  & 65,55 \% & 58,00 \% & \textbf{70,10 \%}\\ [0.75ex]
    \hline
    Average  &  67,53 \% & 67,62 \% &  \textbf{71,57 \%} \\ [0.75ex]
    \hline
    \end{tabular}
    \caption{Results from variant of our models : baseline of BMMN, BMMN-BAE-1 and BMMN-BAE-2.}
    \label{table:final results}
    \end{table}

Based on the total accuracy across affect labels, we can see that the introduction of \emph{z} boosts the accuracy compared to the baseline. Notably when \emph{z} is conflated together following the scheme of BMMN-BAE-2 models with quite a large margin of difference (71.57\% vs 67.62\%),  while the observed improvement of BMMN-BAE-1 is negligible (less than 1\%). In general, for the most considered important emotion dimensions of valence, arousal and liking \cite{Lang}, we found that the introduction of Z improves our models accuracy on both arousal and liking estimates, while we observe slight drop of accuracy for valence.

Specifically, while following the scheme of BMMN-BAE-1 leads to improved results on several affect estimates, such as liking and fear, however in overall, it produces comparable results. This may indicate that our baseline model is capable enough to indirectly captured its own internal features in their affect estimations, though not as compact as \emph{z} used in BMMN-BAE-1 models. In other hand, we found noticeable improvements when \emph{z} is used as complementary input along with raw signal as such BMMN-BAE-2 scheme, with minimal sacrifices in some affect estimates. Further comparison of the results of BMMN with respect to the BMMN-BAE-2, we observe more than 10\% accuracy improvement for anger and disgust emotions, and around 5\% for sadness and neutral which suggests that the \emph{z} may helps to estimates such emotions. Notice that these emotion states require a high level of arousal activation, which may explain close to 10\% accuracy gain (87\% from original 78.36\%) in arousal estimation. 

Based on these findings, we can conclude that the extracted latent features is beneficial on our model estimates, given that it is appropriately integrated. Which in this case is by using it as complementary information along with the raw signals for all modalities.

\subsection{Comparison Against Other Studies}

Table~\ref{table:relative comparison} presents the comparison of our best obtained results from previous sections against other reported results on the AMIGOS \cite{AMIGOS} dataset. We evaluate their results in terms of valence and arousal domain considering their extensive uses in consensus \cite{interdomain} and their availability on all compared studies. In addition, to investigate the importance of physiological signal for affect estimations,  we also add state of the art, deep learning-based affect networks from Kollias et al. \cite{tensorflow}, which only uses facial features, i.e. without any bio-signal input in their emotion estimations pipeline. 



\begin{table}[h!]
\resizebox{\columnwidth}{!}{%
\centering
\renewcommand{\arraystretch}{1.4}
\begin{tabular}{c c cc}
\hline
Methods & Modality & Valence & Arousal \\
\hline
Kollias et al. \cite{tensorflow} &  Faces & 48,76 \% & 60,35 \% \\ [0.25ex]
Gjoreski et al. \cite{interdomain} & ECG, EDA  &- & 56,00 \% \\ [0.25ex]
Yang and Lee \cite{vae} & EEG, ECG, EDA & 68,80 \% & 67,00 \% \\ [0.25ex]
Santamaria et al. \cite{Amigos_1} & ECG, EDA & 76,00 \% & 75,00 \% \\ [0.25ex]
Siddharth et al. \cite{towards_bio} & Faces, EEG, EDA, ECG  & \textbf{83,94 \%} & 82,76 \%  \\ [0.25ex]

BMMN-BAE-2 (ours) & Faces, ECG, EDA & 65,05 \% & \textbf{87,53 \%} \\ [0.25ex]
\hline
\end{tabular}%
}
\caption{Accuracy comparison with other related studies.}
\label{table:relative comparison}
\end{table}



We can summarize several findings from these comparisons. Firstly, our model produces relatively comparable results against other approaches, with highest arousal estimation of 87.53\% of accuracy. Even though it produces lower valence accuracy against the work of Gjoreski et al. \cite{interdomain} and Siddarth et al. \cite{towards_bio}, we need to note that their models require more elaborate EEG features, that demands more extensive instruments. Unlike others, however, our models only need to make efficient use of ECG and EDA signals alongside the face for inference, which is relatively easy to obtain. 

Secondly, we notice that some approaches which do not involve joint modality of faces and bio-signals yield inferior results against other multi-modal approaches, which includes ours. The finding that complies by our conclusion in modality analysis suggesting the importance of multiple modalities input. Another important finding is relatively low results produced by Kollias et al. \cite{tensorflow} that only utilizes the facial features, providing the most moderate valence accuracy. This may be attributed to its lack of any bio-signal modality which may further help their estimations, which explains relatively higher results of other models that exploit them. Finally, we can see in overall a quite big margin of difference between the results of the handcrafted feature-based model, such as Gjoreski et al. \cite{interdomain} to other compared models, which are deep-learning-based approach suggesting the superiority of the latter.







\subsection{Assessment on Anxiety Therapy}
\begin{figure}[h!]
    \centering
    \includegraphics[width=0.75\linewidth]{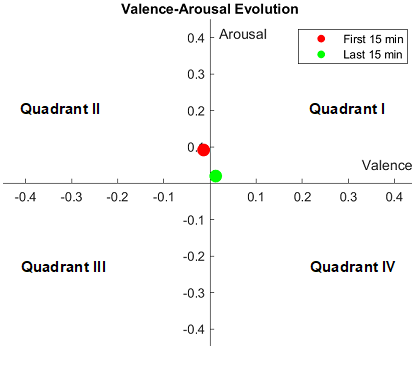}
    \caption{Changes on valence and arousal emotion of all patients, prior (in red) and after the treatment (in green). Scaled from -1 to 1. }
    \label{fig:therapy}
\end{figure}

We use our best performing model of BMMN-BAE-2 to evaluate the anxiety therapy using our collected dataset. Specifically, we run our model on the first 15 minutes subset of the data to obtain the mean emotions states of each patient prior the treatment. Then we collect another estimate for the last 15 minutes to represent the patients condition after the treatment. We accumulate these results for all patients and display them on each of their correspondent quadrant locations as shown on Figure \ref{fig:therapy} to see the corresponding changes.

We can see from the figure that our model able to produce quite sensible results, i.e. it shows the ideal tendency prior and post treatment on emotion changes (from second to four quadrant). We also notice that our models estimates are located close to the centre of each quadrant (neutral emotions). This may caused by our latent features, which are conditioned to specific characteristics of AMIGOS dataset it is trained upon, that we recall in the majority, all samples are located close to the neutral position (see Figure~\ref{fig:amigos_anali}). Thus our training still lacks substantial amount of extreme examples. Nonetheless, our estimates still able to highlight the expected tendency of the treatment, though not too extreme, indicating its real-world capability.

\section{Conclusions}

In this paper, we present a novel multi-modal emotion recognition approach with Internal Auto-Encoder, which operates by extracting features efficiently from several modalities input. We propose a multi-stage pre-processing step to ease our model training on AMIGOS dataset, and we introduce a new therapy dataset to expand the real-world application of our models, and affective computing in general. We create our baseline models by the use of both physiological signals and facial features. Furthermore, we present an improvement by using bio latent features extracted using our internal bio auto-encoder. 

The experiments using AMIGOS dataset provided us several findings: in the first place we observed the importance of multi-modality inputs to achieve higher accuracy compared to individual use of each modality independently. Secondly, we confirmed the benefit of using the latent features, notably by combining it with the original signal, which greatly improved our results from the baseline. Third, the comparison against other related studies in this dataset revealed our competitive results with higher arousal accuracy in overall. We also found that the current state of the art facial based on affect model, which ignores the importance of bio-signal modality, produced lower results than other multi-modal based approach. This finding further supported our conclusion regarding the benefit of multi-modalities input. 

Finally, we applied our best performing models to our collected dataset to identify the emotion changes of the therapy to the patients. We later demonstrate that in fact, our model successfully shows the tendency of the ideal evolution of the patients emotion states, which reflects our models capability on the real-world application. In our future work, we seek to collect more examples to our dataset and adapt our models to its specific characteristics.

\section{ACKNOWLEDGMENTS}
The authors would like to thank the research group of  AMIGOS database used in this research and for granting us access to this dataset. We would also like to show our gratitude to Manel Ballester from Hospital de la Santa Creu i Sant Pau (Barcelona) for his helps on data collections of medical therapy, and Oriol Pujol Martinez for the supports on developing the software for data acquisition.
This work is partly supported by the Spanish Ministry of Economy and Competitiveness under project grant TIN2017-90124-P and the donation bahi2018-19 to the CMTech at the UPF.

\bibliographystyle{unsrt}
\bibliography{sample}

\end{document}